\documentclass[conference]{IEEEtran}
\IEEEoverridecommandlockouts
\usepackage{cite}
\usepackage{amsmath,amssymb,amsfonts}
\usepackage{mathtools}
\newtagform{brackets}{(}{)}
\usetagform{brackets}
\usepackage{algorithm}
\usepackage{algpseudocode}
\usepackage{graphicx}
\usepackage{textcomp}
\usepackage[dvipsnames]{xcolor}
\usepackage{hyperref}
\def\BibTeX{{\rm B\kern-.05em{\sc i\kern-.025em b}\kern-.08em
    T\kern-.1667em\lower.7ex\hbox{E}\kern-.125emX}}

\usepackage{tabularray}
\usepackage{float}

\begin{document}

\title{mhGPT: A Lightweight Generative Pre-Trained Transformer for Mental Health Text Analysis}

\author{\IEEEauthorblockN{Dae-young Kim\textsuperscript{1}, Rebecca Hwa\textsuperscript{2}, Muhammad Mahbubur Rahman\textsuperscript{1,2}\IEEEauthorrefmark{2}}
\IEEEauthorblockA{
   \textsuperscript{1}\textit{Center for Translational Research,}
   \textit{Children’s National Hospital,}
   Washington, DC\\
\{dkim1, mmrahman\}@childrensnational.org}

\IEEEauthorblockA{
   \textsuperscript{2}\textit{George Washington University,}
   Washington, DC\\
   rebecca.hwa@email.gwu.edu}
}

\maketitle
\begin{abstract}
This paper introduces mhGPT, a lightweight generative pre-trained transformer trained on mental health-related social media and PubMed articles. Fine-tuned for specific mental health tasks, mhGPT was evaluated under limited hardware constraints and compared with state-of-the-art models like MentaLLaMA and Gemma. Despite having only 1.98 billion parameters and using just 5\% of the dataset, mhGPT outperformed larger models and matched the performance of models trained on significantly more data. The key contributions include integrating diverse mental health data, creating a custom tokenizer, and optimizing a smaller architecture for low-resource settings. This research could advance AI-driven mental health care, especially in areas with limited computing power.
\end{abstract}

\section{Introduction}

Mental health has a profound impact on the overall quality of life \cite{Berghofer2020Quality, Olatunji2007Quality, Evans2006The}. However, access to mental health services is challenging due to stigma, human resource shortages, fragmented service delivery models, and lack of research capacity for implementation and policy change \cite{wainberg2017challenges}. Meanwhile, Natural Language Processing (NLP) techniques have demonstrated their capability to aid with mental disorders \cite{binggui2022natural}. For example, Jackson et al. developed a sentence classification model that captures key symptoms of Severe Mental Illness (SMI) from the clinical text for the secondary use of mental healthcare data in research \cite{jackson2017natural}. Also, Desouza et al. presented the possibility of NLP to assess Late-Life Depression (LLD) and other comorbidities in elderly demographics \cite{desouza2021natural}. Furthermore, as LLMs came into the spotlight, Dai et al. trained the BERT \cite{devlin2018bert} model with mental health clinical notes and fine-tuned the pre-trained model to classify five major mental disorders \cite{dai2021deep}. However, while the application of NLP or LLM methods demonstrated its promising capability, developing NLP models or training large language models (LLMs) requires high computational resources \cite{qin2021knowledge}, which are often unavailable for many organizations \cite{kim2019deep}. Also, regulations such as the Health Insurance Portability and Accountability Act (HIPAA) and  General Data Protection Regulation (GDPR) hinder the utilization of Infrastructure as a Service (IaaS) to overcome the lack of computing resources. 

\begin{figure*}[ht]
    \centering
    \includegraphics[width=\textwidth]{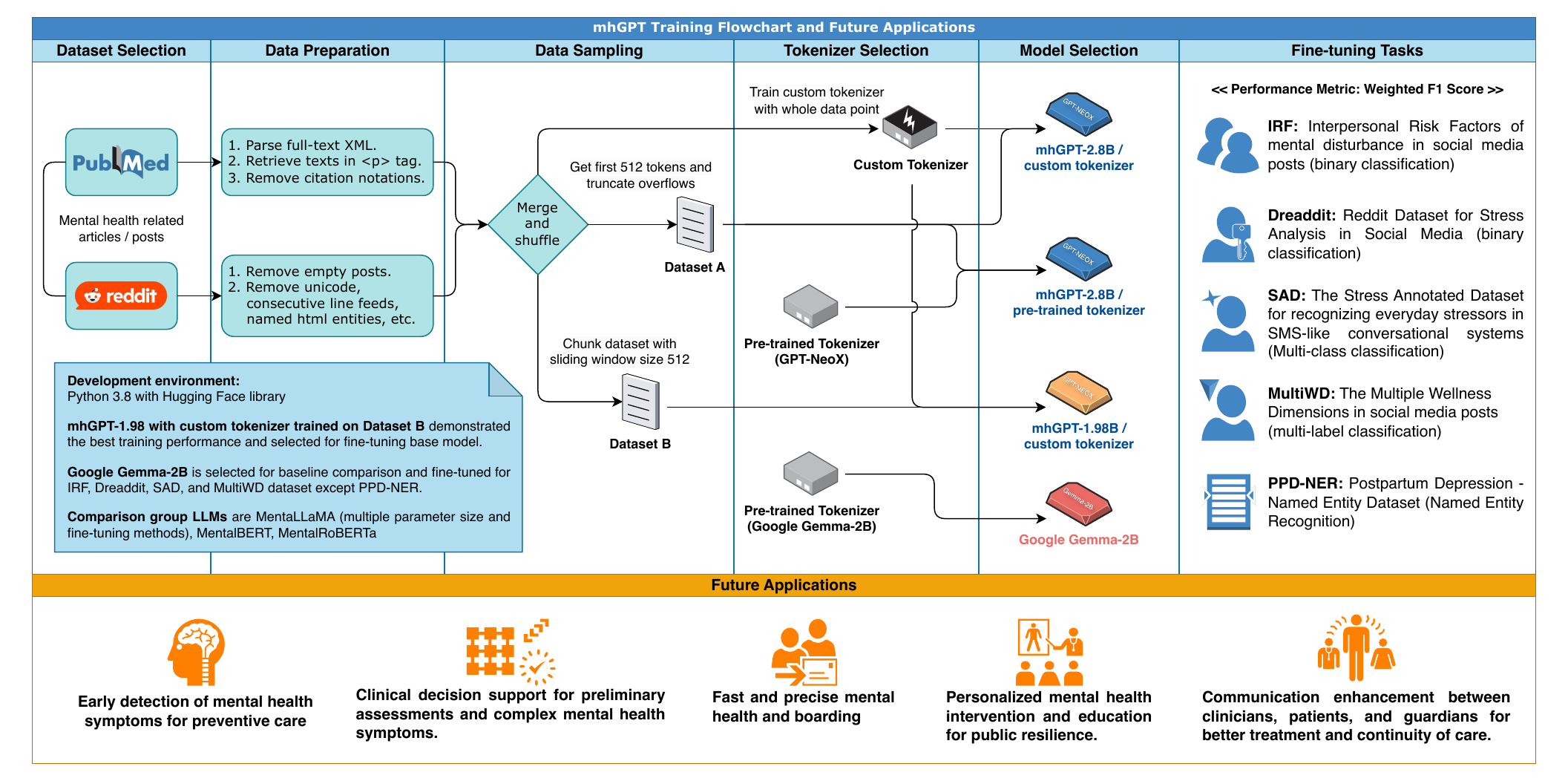}
    \caption{mhGPT Overview}\label{fig:llm-flow}
    \vspace{-1.5em}
\end{figure*}

This paper introduces mhGPT, a high-performance mental health LLM with a small parameter size for low computing resource environments. The training dataset consists of mental health-related PubMed articles and Reddit posts, considering the patient-physician communication perspective: patients express their symptoms in relatively casually structured words due to their lack of professional knowledge, while physicians have to organize mental illnesses or disorders in a formal and professional structure, such as clinical notes and standard test results. This research illustrates decision-making on data preparation and fine-tuning methods as illustrated in figure \ref{fig:llm-flow} while developing mhGPT. Also, we evaluated the performance of mhGPT with state-of-the-art LLMs, such as MentaLLaMA and Google's Gemma.

Our key contributions are as follows. First, this study is the first to integrate research articles from PubMed to train a mental health-specialized LLM. This approach benefits LLMs by allowing them to learn contexts from professional knowledge. Also, this study demonstrated that expert knowledge-infused LLMs with smaller parameter sizes can perform similarly to or even better than the latest state-of-the-art mental health LLMs. Specifically, mhGPT outperformed MentaLLaMA \cite{yang2023mentalllama} with notably larger parameter sizes and training datasets. Lastly, this study explored fine-tuning methods to overcome the structural difference between training data and downstream task data and suggested that NEFTune \cite{jain2023neftune} can enhance the performance of Parameter-Efficient Fine-Tuning (PEFT) even though the authors of NEFTune developed it to enhance the performance of instruction-based fine-tuning.

%To address this gap,
We hypothesized that incorporating mental health-related PubMed articles as expert knowledge and Reddit posts as a casual expression into the training dataset could improve the performance of mental health LLM in a limited hardware resource environment. Secondly, we hypothesized that the expert knowledge infused LLM could provide more precise answers to the more narrowed-down fields of the domains, such as postpartum depression. Lastly, we hypothesized that careful decisions on fine-tuning methods, such as data sampling, tokenizer selection, and regularization, can overcome the overfitting problems caused by a model memorizing small datasets.

This paper introduces mhGPT, a mental health LLM trained on PubMed articles and Reddit posts related to mental health. Our study explores different data preparation options and fine-tuning methods to identify high-performance settings for small-parameter-sized LLMs in low-resource environments as illustrated in figure \ref{fig:llm-flow}.

Our \textbf{key contributions} are as follows. First, this study is the first to integrate research articles from PubMed to train a mental health-specialized LLM. This approach benefits LLMs by allowing them to learn contexts from professional knowledge. We have validated this by fine-tuning mhGPT for the named entity recognition task on the postpartum depression dataset \cite{chowdhuri2019extracting} and demonstrated that mhGPT outperformed MetaMapLite (MMLite) developed by mental health professionals (table \ref{tbl:fine-tuning}). Also, this study demonstrated that expert knowledge-infused LLMs with smaller parameter sizes can perform similarly to or even better than the latest state-of-the-art mental health LLMs. Specifically, mhGPT outperformed MentaLLaMA \cite{yang2023mentalllama} with notably larger parameter sizes and training datasets. Lastly, this study explored fine-tuning methods to overcome the structural difference between training data and downstream task data and suggested that NEFTune \cite{jain2023neftune} can enhance the performance of Parameter-Efficient Fine-Tuning (PEFT) even though the authors of NEFTune developed it to enhance the performance of instruction-based fine-tuning.

\section{Related Work}

\subsection{Mental Health LLMs Trained on Social Media Text}
There are few studies on developed mental health LLMs, and they have trained LLMs predominantly on social media data. For example, MentaLLaMA \cite{yang2023mentalllama} is trained on the comprehensive interpretable mental health instruction (IMHI) dataset sourced from various mental health tasks on social media \cite{ouyang2022training} to automate mental health analysis enhanced zero/few-shot scenarios to improve the quality of explanations. MentalBERT \cite{ji2021mentalbert} model addresses a gap in mental healthcare by enabling early detection of mental disorders and suicidal ideation from social content, facilitating effective social intervention. They outperformed general pre-trained language models in mental disorder detection tasks, demonstrating the value of domain-specific language representations for mental health detection. Dai et al. also pre-trained the BERT \cite{devlin2018bert} model with Brief History (BH) and Physical and Mental status Examination (PME) from EHRs. They fine-tuned the pre-trained binary classification models for five mental disorders - major depressive disorder, schizophrenia, bipolar, minor depressive disorder, and dementia and demonstrated that the models with transferred knowledge from the LLM outperformed models without transferred knowledge \cite{dai2021deep}.

\subsection{Fine-tuning LLMs}
Fine-tuning is still necessary to improve model performance \cite{ding2023parameter} beyond few-shot baselines despite the remarkable zero-shot and few-shot learning performance of the latest LLMs \cite{huang2022large} such as Google's Gemini \cite{team2023gemini} and OpenAI's ChatGPT \cite{achiam2023gpt}. PEFT methods enable modifying a small portion of model parameters and efficient fine-tuning for large models. Among them, Low-Rank Adaptation (LoRA) introduces low-rank decomposition matrices for efficient adaptation of LLMs \cite{hu2021lora}. It is one of the best methods for low-resource environments since it reduces GPU memory usage and training time \cite{hou2022meta}. Furthermore, it is possible to reduce memory usage by QLoRA \cite{dettmers2024qlora}, which introduced 4-bit quantization while backpropagating gradients into LoRA.

\section{LLM Training Method}
\subsection{Dataset}
49,812 PubMed Central (PMC) articles related to mental health were collected from the PMC Open Access repository in BioC format via their API \cite{comeau2019pmc}. This API provides access to full-text research articles for text mining and information retrieval research. We parsed the full-text XML files, extracted the text paragraph-by-paragraph, and shuffled the entire dataset. 

Also, We have collected 1,005,027 submission body text, and 6,036,594 comment sections from twelve subreddits are collected by Reddit API with compliance to their policy \cite{reddit2024api} as illustrated in table \ref{tab:reddit}. To pre-preprocess the data, we removed entries with empty submission body text and content that was too short, with less than five words. Then we removed Unicode, consecutive line feeds, named HTML entities, URLs, decimal character reference, and hexadecimal character reference. As a result, we have included 786,716 submissions and 5,765,005 comments to train our mhGPT.

\begin{table}[]
\centering
\caption{Reddit Data Information}
\label{tab:reddit}
\resizebox{\columnwidth}{!}{%
\begin{tabular}{l|rr}
\multicolumn{1}{c|}{Subreddit} & \multicolumn{1}{c}{\# Submissions} & \multicolumn{1}{c}{\# Comments} \\ \hline
r/ADHD                      & 482,369   & 3,775,167 \\
r/ADHD\_Programmers         & 3,373     & 39,449    \\
r/ADHD\_partners            & 3,588     & 44,167    \\
r/ADHDers                   & 1,512     & 14,539    \\
r/AdultADHDSupportGroup     & 1,292     & 9,005     \\
r/Anxiety                   & 78,952    & 321,096   \\
r/TwoXADHD                  & 5,090     & 67,865    \\
r/adhd\_anxiety             & 7,946     & 52,261    \\
r/adhd\_college             & 671       & 3,113     \\
r/ADHDwomen                 & 54,764    & 543,591   \\
r/Depression                & 225,561   & 510,775   \\
r/SocialAnxiety             & 139,909   & 655,566   \\ \hline
Total Before Pre-processing & 1,005,027 & 6,036,594 \\
Total After Pre-processing  & 786,716   & 5,765,005
\end{tabular}%
}
\end{table}

\begin{figure*}[ht]
    \centering
    \includegraphics[width=0.9\textwidth]{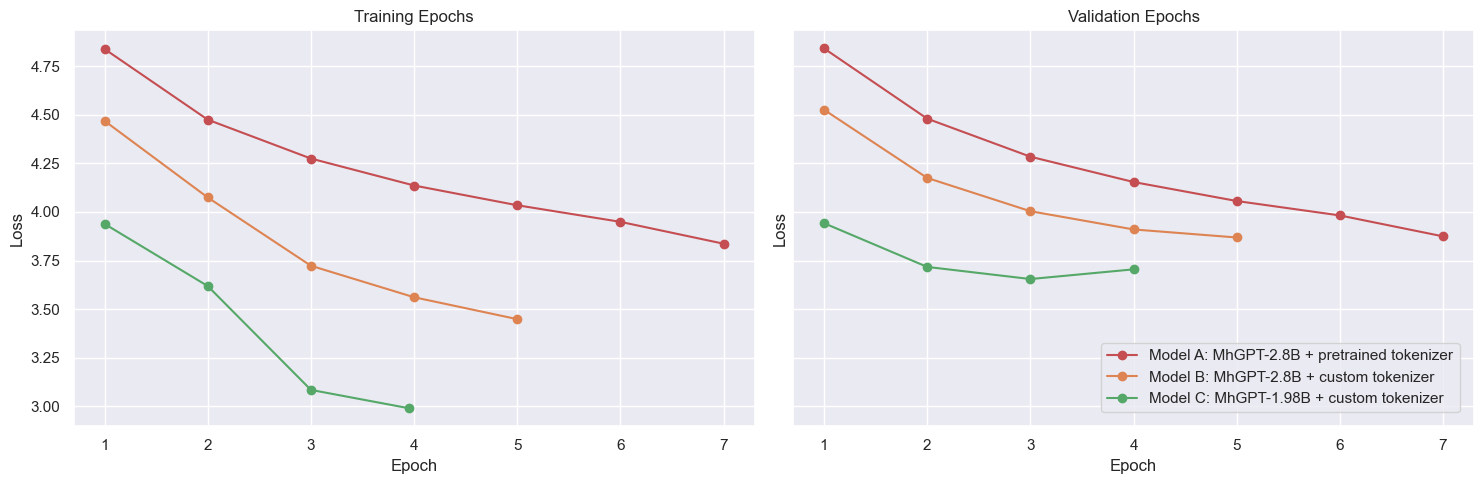}
    \caption{LLM training progress comparison.}\label{fig:llm-training}
    \vspace{-1.5em}
\end{figure*}

\subsection{Data Sampling}
\label{subsub:sampling}
We have sampled our dataset with two approaches for efficient model training on limited environments. First, we set our maximum input length as 512, truncated overflowing tokens, and discarded them. As a result, every row has a maximum length of 512 (\textbf{dataset A}). On the contrary, we did not discard overflowing tokens for the second approach. Instead, we have chunked the dataset using the sliding window method, which has a length 512 and a step size 512. So, the dataset is divided into chunks of identical size. Then, we sampled 5\% from the chunked dataset with a stratified method to maintain PubMed, Reddit submission body text, and comment text ratio (\textbf{dataset B}). The rationale behind adopting sliding window data sampling is that it can increase diversity of continuous data \cite{drosou2009diversity, zhou2022personalized}.

\subsection{Vocabulary}
\label{sub:vocab}
We used two tokenizers and compared fine-tuning performances of mhGPT with each tokenizer. First, we adopted default GPT-NeoX tokenizer (\textbf{pretrained tokenizer}). Secondly, we trained GPT-NeoX tokenizer from the scratch on original full-sized dataset (\textbf{custom tokenizer}) before sampling. The custom tokenizer has the vocabulary size of 52,000 with maximum input length as 512 while keeping pretrained tokenizer's original hyper parameter configuration. %\cite{luo2022biogpt}
% To provide more details, we preserved its original special tokens and byte-level Byte-Pair-Encoding configurations while set maximum input length as 512.
\subsection{Base Model}
\label{sub:model}
The GPT-NeoX model architecture is a base model of the mhGPT. The GPT-NeoX model is an augmented model of NVIDIA's Megatron Language Model (Megatron-LM) and has similar architecture to GPT-3 with new techniques and novel optimizations, released with Apache-2.0 license \cite{Andonian_GPT-NeoX_Large_Scale_2021}. The GPT-NeoX adopted RoPE \cite{su2024roformer} to the first 25\% of embedding vector dimensions for faster convergence and better generalization \cite{wang2021gpt} instead of adding the learned positional embeddings to the token embedding \cite{radford2018improving} used in GPT models. 
% Also, it applied Parallel Attention and Feed forward (PAF) for the better computational efficiency compared to sequential attention methods and to capture wide range of information and nuances. 

We've trained two different parameter size models - 2.8 billion and 1.98 billion parameters. The 2.8B model consists of 22 layers, 3,072 hidden size, 12,288 intermediate feed forward layer size, and 32 attention heads. On the other hand, the 1.98B model has 22 layers, 3,072 hidden size, 6,144 intermediate feed forward layer size, and 64 attention heads.

\subsection{Training Configuration}
\label{sub:train-config}
We set three different configurations for data sampling methods, tokenizers, and model parameter sizes as illustrated in subsections \ref{subsub:sampling}, \ref{sub:vocab}, and \ref{sub:model}. Our final model training configurations are listed below:
\begin{itemize}[noitemsep,topsep=0pt]
    \item \textbf{Model A}: 2.8B parameters, pretrained tokenizer, trained on dataset A.
    \item \textbf{Model B}: 2.8B parameters, custom tokenizer, trained on dataset A.
    \item \textbf{Model C}: 1.98B parameters, custom tokenizer, trained on dataset B.
\end{itemize}

The training configuration of mhGPT is as follows: we set 100 warmup steps with a maximum learning rate of $0.97 \times 10^{-5}$. The weight decay is set to 0.01 to prevent overfitting. The training batch size is configured to occupy all GPU memory except for 1 GB reserved as a buffer for memory usage spikes, aiming for optimal results \cite{mccandlish2018empirical}. The default number of epochs is set to 5 unless the model starts to overfit. 

Models A and B are trained in our organization's High-Performance Computing cluster (HPC), which is configured with NVIDIA H100 GPUs with a maximum VRAM capacity of 60 GB, managed by the Slurm workload manager on Red Hat Enterprise Linux release 8.9 (Ootpa). Model C and all fine-tuning models are trained on an Amazon EC2 instance (g5.12xlarge) consisting of NVIDIA A10G Tensor Core GPUs with a maximum VRAM capacity of 96 GB, using the Deep Learning Proprietary Nvidia Driver AMI GPU PyTorch 2.1.0 Amazon Machine Image (AMI). Average memory usage during the fine-tuning was around 24 GB.
% Overall, no budgets were used to utilize HPC, and \$3223 was used to train model C and fine-tune all downstream task models in the AWS EC2 instance. 

For data preparations and model trainings, Python 3.8 and following packages with compatible versions are used: Pandas \cite{reback2020pandas, mckinney-proc-scipy-2010}, Scikit-learn \cite{pedregosa2011scikit}, Spacy \cite{Honnibal_spaCy_Industrial-strength_Natural_2020}, Hugging Face Datasets \cite{lhoest-etal-2021-datasets}, and Hugging Face Transformers \cite{wolf2019huggingface}.

\subsection{Model Selection}
We compared the training performance of three models, A, B, and C, based on different tokenizers and data sampling methods.

First, we compared the training performance of models A and B to compare model training performance based on different tokenizer configurations. As illustrated in figure \ref{fig:llm-training}, model B's validation loss in epoch five was smaller than that of model A in epoch seven. Then, we compared the training performance of models B and C to compare model training performance based on data sample methods. Model C has shown better performance on validation loss. Specifically, it was trained on a 5\% sampled dataset, tokenized by a custom tokenizer, and chunked using a sliding window method. This approach allowed for a more comprehensive representation of the dataset, as the randomly selected chunks from the entire dataset exhibited a greater diversity in structure and content compared to the sum of the first 512 tokens of each data point.

\section{Fine-Tuning Method}
\label{sec:ftmethod}
\subsection{Fine-Tuning Configurations}
We adopted PEFT method to fine-tune mhGPT. The PEFT fine-tuning processes are done as follows. First, a dataset is tokenized by our custom GPT-NeoX tokenizer trained on LLM training dataset. Secondly, a task-specific layers are added on top of mhGPT. Then, newly added layers and part of a pretrained mhGPT parameters are trained on a down-stream task dataset according to PEFT configurations. In particular, we adopted LoRA \cite{hu2021lora} because of its advantages on reducing GPU memory usage and training time \cite{hou2022meta}. We targeted linear layers only with the LoRA configurations of rank 64 and alpha 32. To further reduce GPU memory usage, we adopted 4-bit quantization with QLoRA method \cite{liu2023llm, dettmers2024qlora}. 

One challenge of the LoRA method with 4-bit quantization was rapid overfitting. To address this issue, we incorporated NEFTune \cite{jain2023neftune} during training (\ref{sub:neftune}) because the combination of LoRA and NEFTune proved to be effective in maintaining model performance without compromising generalization. Lastly, Default leraning rate is set to 0.9e-5 since mhGPT is trained with learning rate 0.97e-5. In the following sections, fine-tuned models are trained with these default configurations if no changes are specified. 

In most cases, the default hyperparameters illustrated above worked well during the fine-tuning process. However, when we had to change values of the hyperparameters to improve performance, we set and followed a ``regulation over complexity'' criteria. The rationale behind the criteria is that mhGPT already knows how to understand input texts and all we have to do is to less train as possible to perform downstream tasks to make sure that it applies pre-trained knowledge to downstream tasks, rather than it fits parameters to the dataset. The second important criteria was to be aware of consequent trade-offs hyperparameters value change.

\subsection{NEFTune Application to PEFT}
\label{sub:neftune}
We adopted NEFTune noise embeddings to enhance PEFT fine-tuning performance, especially to overcome overfitting and imbalanced datasets. We adopted NEFTune because it adds noise to the data but does not change the semantic relationship between tokens \cite{jain2023neftune}. Though the primary purpose of the method is to improve the performance of instruction-based fine-tuning, we focused on the fact that the method adds noise to the tokens in the vocabulary. Then, we hypothesized that NEFTune could improve PEFT fine-tuning performance since PEFT and instruction-based methods share the same data input formats. Our default NEFTune noise alpha value is ten because the number resulted in the best performances in its original paper.

\begin{figure}[t]
    \centering
    \includegraphics[width=\columnwidth]{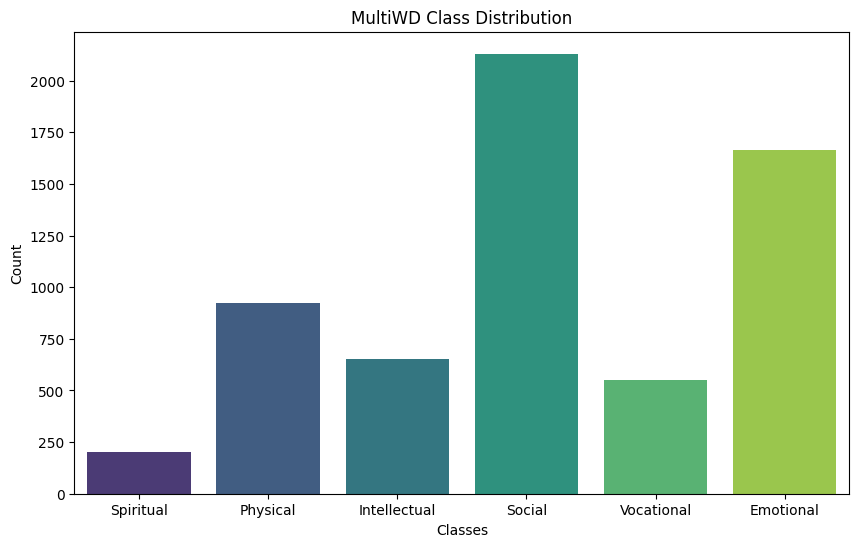}
    \caption{MultiWD dataset class distribution.}\label{fig:multiwd-dist}
    \vspace{-1.5em}
\end{figure}

\subsection{Baseline Comparison}
We fine-tuned Google Gemma-2B for each task as a baseline performace and evaluate fine-tuning performance of mhGPT. The LoRA hyperparameters are the same with mhGPT fine-tuning configuration and maximum learning rate is set to 2e-5.

\begin{figure*}[t]
    \centering
    \includegraphics[width=0.9\textwidth]{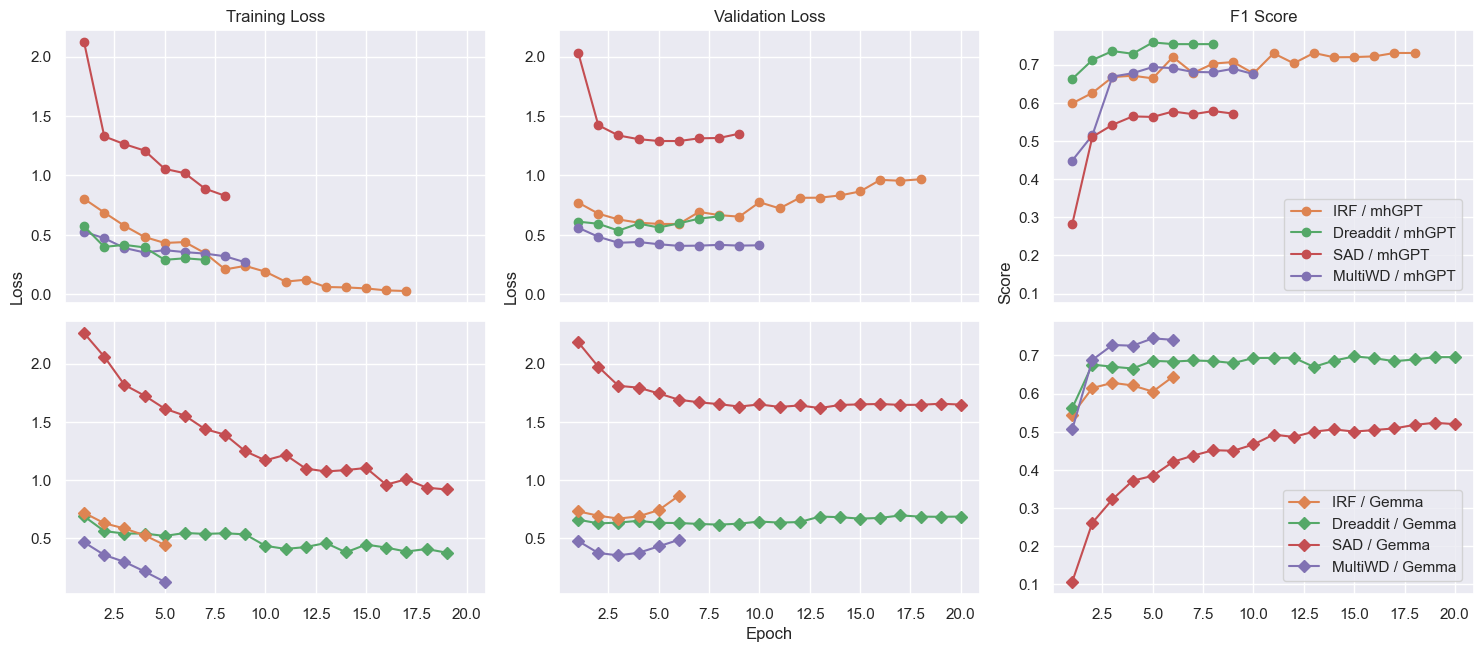}
    \caption{mhGPT-1.98B fine-tuning comparison with baseline model Google Gemma-2B.}
    \label{fig:fine-baseline}
    \vspace{-1.5em}
\end{figure*}

\section{Fine-Tuning Datasets}
In this section, we explain datasets that are used to fine-tune and evaluate downstream models of the pre-treained mhGPT. We fine-tuned mhGPT on five datasets for four downstream tasks.

\subsection{Binary Sequence Classification}
We utilized an annotated dataset for explainable Interpersonal Risk Factors of mental disturbance in social media posts (\textbf{IRF}) \cite{garg2023annotated} and the Reddit Dataset for Stress Analysis in Social Media (\textbf{Dreaddit}) \cite{elsbeth2019dreaddit} for our binary classification experiments.

The IRF \cite{garg2023annotated} dataset is dataset with human-labelled explanations of interpersonal risk factors that affects mental disturbance on social media, which are \textit{Thwarted Belongingness} (TBE), and \textit{Perceived Burdensomeness} (PBU). The author of the dataset collected 3,522 posts from Reddit \textit{r/depression} and \textit{r/SuicideWatch} subreddits and developed binary classification models for each label. Since their best F1 score was from TBE classification model \cite{yang2023mentalllama}. The classes ``TBE'' and ``No TBE'' are fairly evenly distributed.

The Dreaddit \cite{elsbeth2019dreaddit} is a dataset that identified percieved stress of authors from their Reddit posts. It is a comprehensive collection of social media posts from Reddit across five categories, totaling 190,000 posts. Researchers have manually labeled 3,500 segments from 3,000 posts to facilitate the identification of stress using supervised learning methods. The dataset explored stress across diverse online contexts, addressing a gap in existing research that often focuses on specific domains or short-form platforms. 

\subsection{Multi-Class Sequence Classification}
We used the Stress Annotated Dataset for recognizing everyday stressors in SMS-like conversational systems (\textbf{SAD}) \cite{mauriello2021sad} to evaluate the model's multi-class sequence classification performance. The Stress Annotated Dataset for Recognizing Everyday Stressors in SMS-like Conversational Systems (SAD) is a collection of 6,850 SMS-like sentences categorized into nine stressor categories. These categories were developed from stress management literature, live chatbot conversations, crowdsourcing, and targeted web scraping. The purpose of the dataset is to enable chatbots to classify stress inputs to provide appropriate advice during conversations. The NEFTune alpha is set to 5 because the fine-tuning model did not fit well on the training dataset.

\begin{table*}[h]
\centering
\caption{Fine-tuning Performance Comparison}
\resizebox{0.8\textwidth}{!}{
\begin{tblr}{
  column{2} = {c},
  column{3} = {c},
  column{4} = {c},
  column{5} = {c},
  cell{1}{1} = {c},
  cell{2}{1} = {r=10}{c},
  cell{2}{2} = {r=5}{},
  cell{7}{2} = {r=5}{c},
  cell{12}{1} = {r=5}{c},
  cell{12}{2} = {r=5}{},
  cell{17}{1} = {r=5}{c},
  cell{17}{2} = {r=5}{},
  cell{22}{1} = {r=2}{c},
  cell{22}{2} = {r=2}{},
  cell{23}{3} = {c=2}{},
  vline{1-3,5-6} = {1-23}{},
  hline{1,12,17,22,24} = {-}{},
  hline{2} = {1-5}{},
  hline{7} = {1-5}{},
}
\textbf{Task}                       & \textbf{Dataset}  & \textbf{Base Model}                                       & \textbf{Fine-tuning Method} & \textbf{F1}             \\
{Binary\\Classification}            & IRF               & mhGPT-1.98B                                               & PEFT - LoRA                 & 71.99                   \\
                                    &                   & MentaLLaMA-7B \cite{yang2023mentalllama}                  & Instruction-tuning          & 67.53                   \\
                                    &                   & MentaLLaMA-chat-7B \cite{yang2023mentalllama}             & Instruction-tuning          & 72.88                   \\
                                    &                   & MentaLLaMA-chat-13B \cite{yang2023mentalllama}            & Instruction-tuning          & 76.49                   \\
                                    &                   & Gemma-2B \cite{team2024gemma}                             & PEFT - LoRA                 & \textbf{81.22}          \\
                                    & Dreaddit          & mhGPT-1.98B                                               & PEFT - LoRA                 & 73.55                   \\
                                    &                   & MentaLLaMA-7B                                             & Instruction-tuning          & 71.65                   \\
                                    &                   & MentaLLaMA-chat-7B                                        & Instruction-tuning          & 62.20                   \\
                                    &                   & MentaLLaMA-chat-13B                                       & Instruction-tuning          & \textbf{75.79}          \\
                                    &                   & Gemma-2B                                                  & PEFT - LoRA                 & 68.66                   \\
{Multi-Class\\Classification}       & SAD               & mhGPT-1.98B                                               & PEFT - LoRA                 & 57.71                   \\
                                    &                   & MentaLLaMA-7B                                             & Instruction-tuning          & 49.93                   \\
                                    &                   & MentaLLaMA-chat-7B                                        & Instruction-tuning          & 62.18                   \\
                                    &                   & MentaLLaMA-chat-13B                                       & Instruction-tuning          & \textbf{63.62}          \\
                                    &                   & Gemma-2B                                                  & PEFT - LoRA                 & 45.16                   \\
{Multi-Label    \\Classification}   & MultiWD           & mhGPT-1.98B                                               & PEFT - LoRA                 & 70.34                   \\
                                    &                   & MentaLLaMA-7B                                             & Instruction-tuning          & 68.44                   \\
                                    &                   & MentaLLaMA-chat-7B                                        & Instruction-tuning          & \textbf{75.79}          \\
                                    &                   & MentaLLaMA-chat-13B                                       & Instruction-tuning          & 75.11                   \\
                                    &                   & Gemma-2B                                                  & PEFT - LoRA                 & 72.73                   \\
{Named Entity \\Recognition}        & PPD-NER           & mhGPT-1.98B                                               & PEFT - LoRA                 & \textbf{88.04}          \\
                                    &                   & MMLite \cite{demner2017metamap,chowdhuri2019extracting}   &                             & 84.36                                    
\end{tblr}
}
\label{tbl:fine-tuning}
\end{table*}

\subsection{Multi-Label Sequence Classification}
We used the Multiple Wellness Dimensions in social media posts (\textbf{MultiWD}) \cite{sathvik2023multiwd} dataset to fine-tune and evaluate the multi-label classification models. Multiple Wellness Dimensions in Social Media Posts (MultiWD) comprises 3,281 instances, serves as a curated collection specifically designed and annotated for identifying wellness dimensions from Reddit posts, with 6 labels: \textit{``Emotional,'' ``Intellectual,'' ``Physical,'' ``Social,'' ``Spiritual,'' ``Vocational.''} As illustratead in figure \ref{fig:multiwd-dist}, the data set has severly unbalanced class distribution. Therefore, NEFTune noise alpha is set to 20 for an attempt to overcome highly imbalanced characteristic of the dataset.

\subsection{Named Entity Recognition}
We used Postpartum Depression Named Entity Dataset (\textbf{PPD-NER}) \cite{chowdhuri2019extracting} dataset to fine-tune and evaluate the NER task model. PPD-NER \cite{chowdhuri2019extracting} is a NER dataset, Postpartum Depression (PPD) related terms annotated on 10,584 forum threads about PPD and its preliminary assessment of topics from \textit{BabyCenter.com} online health communities. The authors of the developed the MetaMapLite (MMLite) \cite{demner2017metamap} that annotate PPD related terms based on the Human Phenotype Ontology (HPO) \cite{kohler2014human, groza2015human} concept recognition software to identify biomedical terms. The dataset provides Fielded MetaMap Indexing file (MMI) \cite{aronson1997mmi} for each thread, which include UMLS preferred name, actual text mapped to the UMLS concept, and positional information in the original thread file.

\section{Result}
\label{sec:result}
We reported the best performance metric score before any possible overfitting occurred. Figure \ref{fig:fine-baseline} describes the training loss, validation loss, and weighted average F1 score of mhGPT and the baseline during the fine-tunings. Table \ref{tbl:fine-tuning} presents a summary of the fine-tuning performance, as well as a comparison with our baseline and the previously reported performance of state-of-the-art LLMs specialized in mental health. 

An initial hypothesis of this project was that LLMs trained on expert knowledge-infused mental health data could perform similarly to or even better than the comparison group of LLMs trained on social media about mental health. Table \ref{tbl:fine-tuning} illustrates that mhGPT outperformed at least one variation of MentaLLaMA for every task and dataset, and it showed similar performance compared to MentalBERT and MentalRoBERTa. Compared to mhGPT, all three models were trained on only social media data and had much higher parameterized architecture.

It was also hypothesized that the expert knowledge-infused LLM could provide more precise answers to subdomain fields of mental health. This research compared the NER task performance of the mhGPT and MMLite \cite{demner2017metamap} from the PPD-NER dataset's original research paper \cite{chowdhuri2019extracting}. The authors of the paper reported recall and precision scores only, so we derived their F1 score by calculating the harmonic mean of the recall and precision scores. The result showed that mhGPT also outperformed human annotators in the dataset, which indicates that the hypothesis is true.

However, the baseline Gemma-2B also performed well, particularly in the binary classification task with the IRF dataset and the multi-label classification task with the MultiWD dataset. In the case of IRF binary classification, Gemma-2B achieved the best score. We conjecture that this result is due to the substantial size difference in the training datasets. The Gemma-2B model's actual parameter size is 2.51B and the model is trained on a total of 6 trillion tokens from a wide variety of sources such as web documents, programming languages, and mathematics \cite{team2024gemma}. Additionally, we hypothesize that although Gemma-2B has good performance, it may have less interpretability in its decisions since it was not trained on mental health domain-specific data. However, this is beyond the scope of this research and will be explored in subsequent studies.

The third question in the study was whether adopting an appropriate fine-tuning method can overcome the structural difference between the training dataset and downstream task data. We found out that NEFTune \cite{jain2023neftune} can enhance the performance of PEFT fine-tuning on small-sized and imbalanced datasets. The authors of NEFTune evaluated the method on an instruction dataset with 7B parameter LLMs—LLaMA-1 \cite{touvron2023llama}, LLaMA-2 \cite{touvron2023llama2}, and OPT-6.7B \cite{zhang2022opt}. The best NEFTune alpha value of 10 from their research worked well with the plain labeled dataset we used in the much smaller 1.98B parameter model with a different architecture, GPT-NeoX. Additionally, it was possible to regularize highly imbalanced dataset MultiWD (figure \ref{fig:multiwd-dist}). By increasing the NEFTune alpha value to 20, the mhGPT fine-tuned model outperformed MentaLLaMA-7B, which has more than three times the number of parameters.

We also fine-tuned Model A and Model B. The best F1 scores for Model A's fine-tuning on the MultiWD multi-label sequence classification and IRF binary sequence classification tasks were 46.52 and 49.30, respectively. For Model B, the scores were 49.32 and 49.30, respectively. The other two datasets, Dreaddit and SAD, were added later to provide a firm comparison for the mhGPT-1.98B model. 

\section{Conclusion}
In this paper, we introduced mhGPT, a lightweight generative pre-trained transformer for mental health text analysis. mhGPT is based on the GPT-NeoX architecture and was trained using social media and PubMed articles related to mental health. Evaluations are done by fine-tuning on five mental health-related downstream tasks using the IRF, Dreaddit, SAD, MultiWD, and PPD-NER datasets. The fine-tuning performance was compared with current state-of-the-art mental health-specific language models and a general-purpose language model. The results demonstrated that mhGPT outperformed MentaLLaMA and showed similar performance to other LLMs such as MentalBERT and MentalRoBERTa, which have notably larger parameter sizes and training datasets.

Our findings indicate that it is possible to enhance domain-specific LLM's performance by incorporating expert knowledge-infused dataset, custom tokenizer, and sliding window data sampling. Also, we found out that it is possible to enhance PEFT peformance and overcome imbalanced dataset issue by leveraging NEFTune noise embeddings.

\section{Limitations}
% Add more limintations
Despite the promising results demonstrated by mhGPT, our research has limitations. First, we fine-tuned mhGPT for only four downstream tasks. Second, a more comprehensive exploration of interpretability is needed, as any potential decision-making failure in the mental health domain could have severe implications for patients. Third, mhGPT has not been tested by mental health professionals for its validation in real-world settings. Fourth, the social media dataset used for LLM training may have misinformation and disinformation. We did not verify these data, and there is a risk that the model may produce biased output.

\section{Ethical Considerations}
We understand the responsibility of computing professionals and pursue the public good through our research. To avoid any harm and respect privacy, we excluded personally identifiable information (PII) from Reddit posts and PubMed articles on mental health, such as user IDs or information about authors, during the data preparation for LLM training and downstream tasks fine-tuning.

\appendix

\section{Computational Cost}
\label{apx:cost}
We spent \$3,187.83 on our LLM training on AWS using on-demand instances. To get the job done, we used different types of GPU instances, such as p3, g5, trn1, gr6, and g6, for a total of 534.839 hours. We also used Local HPC for LLM training.

\bibliographystyle{IEEEtran}
\bibliography{IEEEabrv, reference.bib}

\end{document}